\documentclass[numbered]{trbunofficial}
\usepackage{amsmath}
\usepackage{graphicx}
\usepackage{float}

\begin{document}
\nolinenumbers

\title{A Two-Stage Hybrid Framework for Inferring Air Trips from Mobile Location-based Service (LBS) Data under Incomplete Airport Sightings}

\TRBauthor{Hsuan-Chih Wang}{Graduate Research Assistant\\Department of Civil and Environmental Engineering, University of Maryland}{hcwang@umd.edu}[College Park, MD, 20742][0000-0002-8669-7915]

\TRBauthor{Xianfeng Terry Yang, Ph.D.}{Associate Professor \& Director \\Maryland Transportation Institute \\Department of Civil and Environmental Engineering, University of Maryland}{xtyang@umd.edu}[College Park, MD, 20742][0000-0002-9416-6882]
\AuthorHeaders{Wang and Yang}

\maketitle
\noindent\textit{Accepted for presentation at the 106th Annual Meeting of the Transportation Research Board, January 2027.}

\section{Abstract}
\noindent\textbf{Objectives:}
Mobile device location-based service (LBS) data provides large-scale spatiotemporal observations that can help airport planners understand air travelers' mobility patterns. However, airport sightings are often missing before or after flights, making strictly rule-based detection incomplete. This study develops a two-stage framework to infer U.S. domestic air trips from large-scale LBS data under incomplete airport sightings. 

\hfill\break%
\noindent\textbf{Methods:}
The proposed framework consists of two stages. Stage~1 uses heuristic rules to identify high-confidence air trips whose endpoints are observed at two distinct airports. Stage~2 addresses latent air trips with a probabilistic model that infers air routes using spatial proximity, temporal compatibility, and historical flight schedule information. Using January 2022 nationwide LBS data with nearly 10 million users and 131 million sightings, the results are validated against T-100 Domestic Segment route benchmarks and the Stage~1 pseudo-ground-truth reference set.

\hfill\break%
\noindent\textbf{Findings:}
The proposed framework increases air route coverage from 10.65\% under Stage~1 to 50.60\%, and recovers nearly all top-1,000 benchmark routes. It improves Spearman rank correlation with T-100 route shares from 0.4490 to 0.7630 and reduces Jensen-Shannon divergence from 0.2808 to 0.2263. Stage~2 also recovers 92.2\% of Stage~1 reference trips while adding broader coverage for semi-observed and latent air trips.

\hfill\break%
\noindent\textbf{Novelty:}
This study formalizes three endpoint observation regimes for LBS data-based air trip inference: airport-observed, semi-observed, and latent cases. It combines rule-based detection with probabilistic inference rather than relying solely on airport sightings or simple kinematic thresholds.

\hfill\break%
\noindent\textbf{Practical Applications:}
The framework helps airport agencies and planners identify domestic air travel patterns that may be missed when airport sightings are incomplete. The inferred air trips can also support analyses of airport access, egress, and route-level demand.

\newpage
\section{Introduction}\label{sec:introduction}

The study of air traveler behaviors is important for both airport planning and broader regional transportation analysis. Air trips are embedded in door-to-door travel chains that connect homes, workplaces, airports, flights, and post-arrival destinations. Understanding these linked movements can help agencies and airport planners evaluate airport access demand, parking operations, multimodal connections, and route-level passenger mobility. Analyzing and forecasting these patterns requires reliable, high-resolution transportation data that can capture travelers' movements beyond the airport terminal. 

Traditional transportation data sources provide only partial visibility into these mobility patterns. Conventional aviation demand studies commonly operate at national, market, or airport scales and rely on socioeconomic, demographic, and airline operational variables \cite{Wan21b,Ban20}. These aggregate data are valuable for forecasting passenger volumes and comparing airport markets, but they generally do not reveal the full door-to-door travel chain of individual passengers at the urban transportation network level. Passenger surveys can provide richer information on traveler characteristics, trip purposes, airport choices, and access modes; however, they are expensive to practitioners, conducted infrequently, and subject to sampling and recall biases. In addition, survey samples are often too small or spatially coarse to support detailed analysis of airport access, egress, and air passenger movements across a national airport network. These limitations motivate the use of large-scale passive mobility data to complement traditional aviation datasets.

On the other hand, mobile location-based service (LBS) data, collected from GPS-enabled mobile devices and app-based location sightings, provides a complementary source for filling these gaps. By capturing passive location sightings from large user populations, LBS data can help reveal where travelers come from before airport arrival, where they go after landing, and how observed mobility traces correspond to air-route demand. However, mobile devices may lose connectivity, be switched to airplane mode during a flight, or be activated only after a traveler has landed at the airport or shortly before arriving at the airport. Heuristic rules are useful for high-confidence air trip detection but are insufficient when airport endpoints are not explicitly sighted, even when the observed movement is physically plausible as an air trip. Such sparse and irregular sightings make air trip inference challenging.

\begin{figure}[h!]
 \centering
 \includegraphics[width=\textwidth]{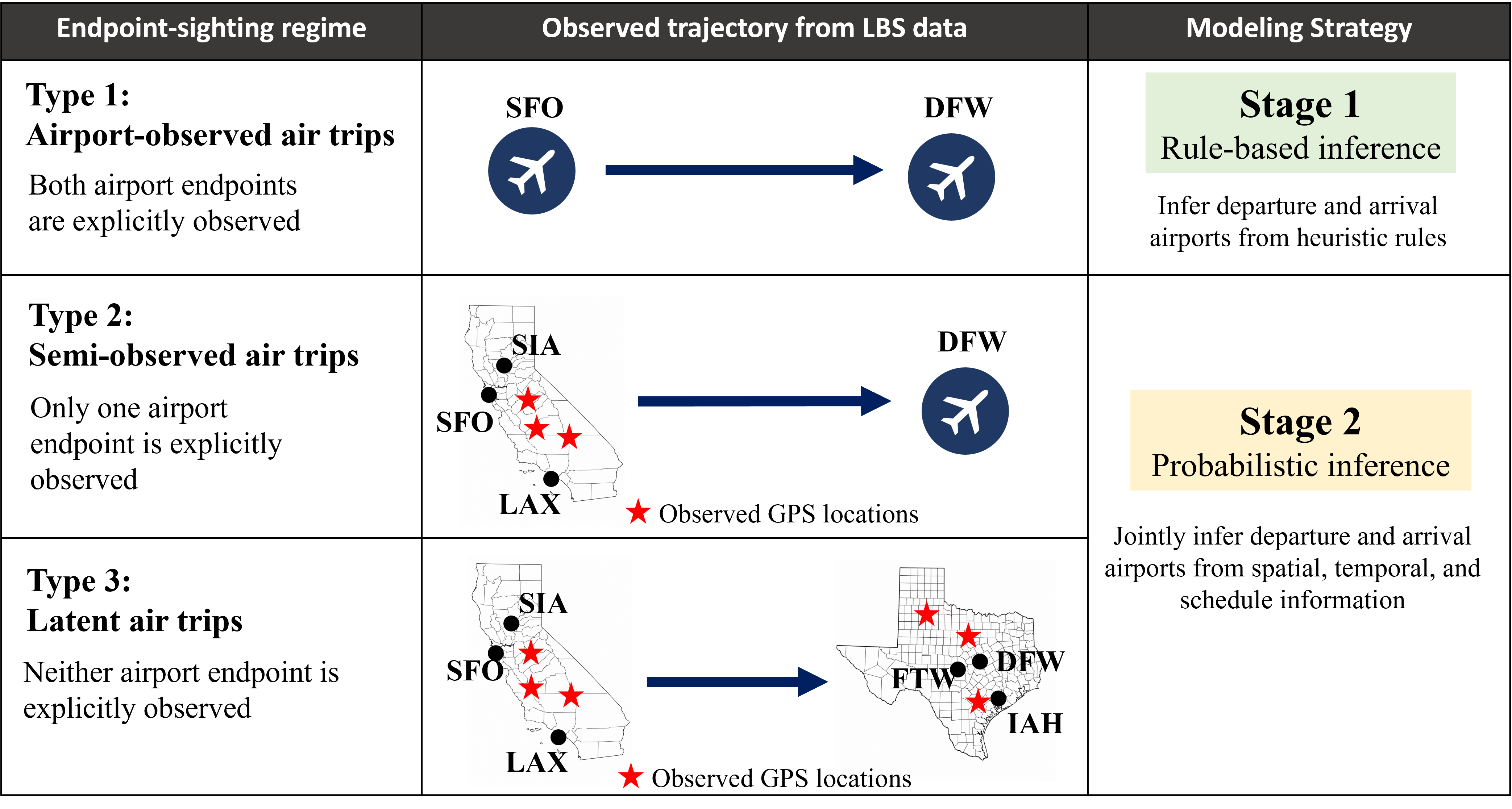}
 \caption{Three endpoint-sighting regimes in LBS-based air-trip inference and their corresponding modeling strategies.}
 \label{fig:observation-regimes}
\end{figure}

To tackle this problem, as shown in Figure~\ref{fig:observation-regimes}, we frame this problem using three endpoint-sighting regimes that require different levels of inference. In Type~1 (airport-observed air trips), the device is sighted at both the departure and arrival airports, supporting strict, high-precision identification. In Type~2 (semi-observed air trips), only one airport endpoint is explicitly sighted; the other endpoint is represented by a non-airport location, which may correspond to a user after leaving the airport or before arriving at the airport. In Type~3 (latent air trips), neither endpoint is explicitly sighted at an airport. Type~2 and Type~3 cases therefore require an inference framework to identify flights that can explain the long spatiotemporal gap between available sightings. This distinction motivates this study: a deterministic filter can reliably identify some Type~1 trips but cannot provide broad coverage across all three regimes. 

To address the above research question, this study develops a two-stage hybrid framework for inferring U.S. domestic air trips from massive LBS data. Stage~1 applies heuristic kinematic rules to identify a high-precision reference set of Type~1 air trips. Stage~2 uses a probabilistic model to infer air routes from spatial, temporal, and historical flight-schedule information for Type~2 and Type~3 air travelers. The framework does not require terminal sensing infrastructure or a standardized airport point-of-interest taxonomy; instead, airport labels are used only as one cue during candidate air route generation and are spatially consolidated before inference.

The contributions of this study are twofold. First, we formalize the airport-observed, semi-observed, and latent air-trip regimes that arise when LBS data sightings are used to infer air travel. Second, we combine a rule-based stage with a probabilistic inference stage, allowing the framework to extend beyond air trips whose endpoints are both explicitly observed at airports. Using a nationwide January 2022 LBS dataset with nearly 10 million users and 131 million raw sightings across the U.S., the analysis demonstrates the framework's capability to recover air routes even when users' sightings are not explicitly recorded at airports. 

The remainder of this paper is organized as follows. First, the literature on aviation demand analysis, LBS data, and air-travel mode classification is reviewed. Next, the study dataset and proposed framework are described. The evaluation and validation results are then presented, followed by conclusions and directions for future research. 

\section{Literature Review}\label{sec:literature}

LBS data have become a major source for observing population mobility at large spatial and temporal scales \cite{Che16,Too15,Kabiri24}. These data include several forms of passively collected location records, including in-vehicle GPS traces, cellular call detail records (CDRs), cellular sighting records, and location-based service (LBS) data. GPS traces can provide high-frequency vehicle locations and have been used for applications such as freight traffic analysis \cite{Cha18}. CDRs and cellular sightings are generated through mobile-network interactions and typically locate devices through cell towers or triangulated network signals rather than precise GPS coordinates \cite{Che16}.

LBS data are commonly collected through smartphone applications using GPS, cellular towers, WiFi, and other connectivity sources \cite{Yang20thesis}. Compared with CDRs and cellular sightings, LBS data can provide finer spatial detail, but they also require careful preprocessing, validation, and privacy protection. In this study, LBS data refers primarily to location sightings that contain anonymized device identifiers, timestamps, coordinates, and place information generated by the data providers.

Traditional aviation demand research has emphasized aggregate forecasting at national, market, and airport levels \cite{Wan21b,Ban20}. These studies are useful for demand prediction, but they provide limited visibility into individual travel chains connecting homes, workplaces, airports, and flights. Surveys and terminal sensing can add detail, but surveys are episodic and costly, while WiFi, Bluetooth, and other local sensing systems usually lack network-wide coverage \cite{Bur22,Sch14,Kur17,Mil23}.

Recent aviation studies have shown that LBS data can support airport and air-passenger analysis by measuring airport catchment areas, passenger flows, access and egress patterns, connection volumes, and airport activity patterns \cite{Mar19,Bur22}. These applications demonstrate that passive mobility traces can extend observation beyond the terminal and provide a broader view of how travelers interact with the aviation system. 

LBS data has also been widely used for origin-destination estimation and travel mode inference \cite{Che16,Too15,Bac19b,Yan20b}. Studies for transportation mode classification have shown substantial progress for ground transportation modes, including car, walk, bicycle, bus, rail, subway, and tram, especially when GPS or cellular observations are frequent enough to support feature extraction from trajectories \cite{Bol12,Chi19,Hua19b,Sad21,Yan18b}. These studies provide important methodological foundations for working with sparse and noisy mobility traces, but their assumptions do not fully transfer to air trips, where long observation gaps, variable airport dwell times, and missing endpoint sightings are common.

Consequently, existing LBS-based air-trip classification methods often rely on heuristic thresholds for travel distance, travel duration, speed, and airport proximity \cite{Yan20b,Bol12,Chi19,Hua19b,Sad21}. These rules can identify clear airport-sighted air trips with high precision, but they are less effective for route inference when one or both airport endpoints are absent from the trace. In semi-observed and latent cases, multiple airport pairs may be compatible with the same long-distance spatiotemporal gap, particularly in multi-airport regions or along dense domestic air corridors. This limitation motivates a probabilistic air trip inference approach that integrates spatial evidence, temporal feasibility, and historical flight service rather than relying solely on deterministic thresholds.

Overall, the reviewed literature indicates that LBS data are valuable for aviation analysis, but existing studies do not fully address air trip inference under incomplete endpoint sightings. A scalable framework is needed to distinguish airport-observed, semi-observed, and latent air-trip regimes; infer directed domestic air trips from sparse sightings; and evaluate inferred air routes independently of the constraints used during inference. This study addresses this gap through a two-stage framework that combines rule-based air-trip detection, probabilistic air-trip inference, and validation based on both internal consistency and external route-level benchmarks.

\begin{figure}[!t]
 \centering
 \includegraphics[width=\textwidth]{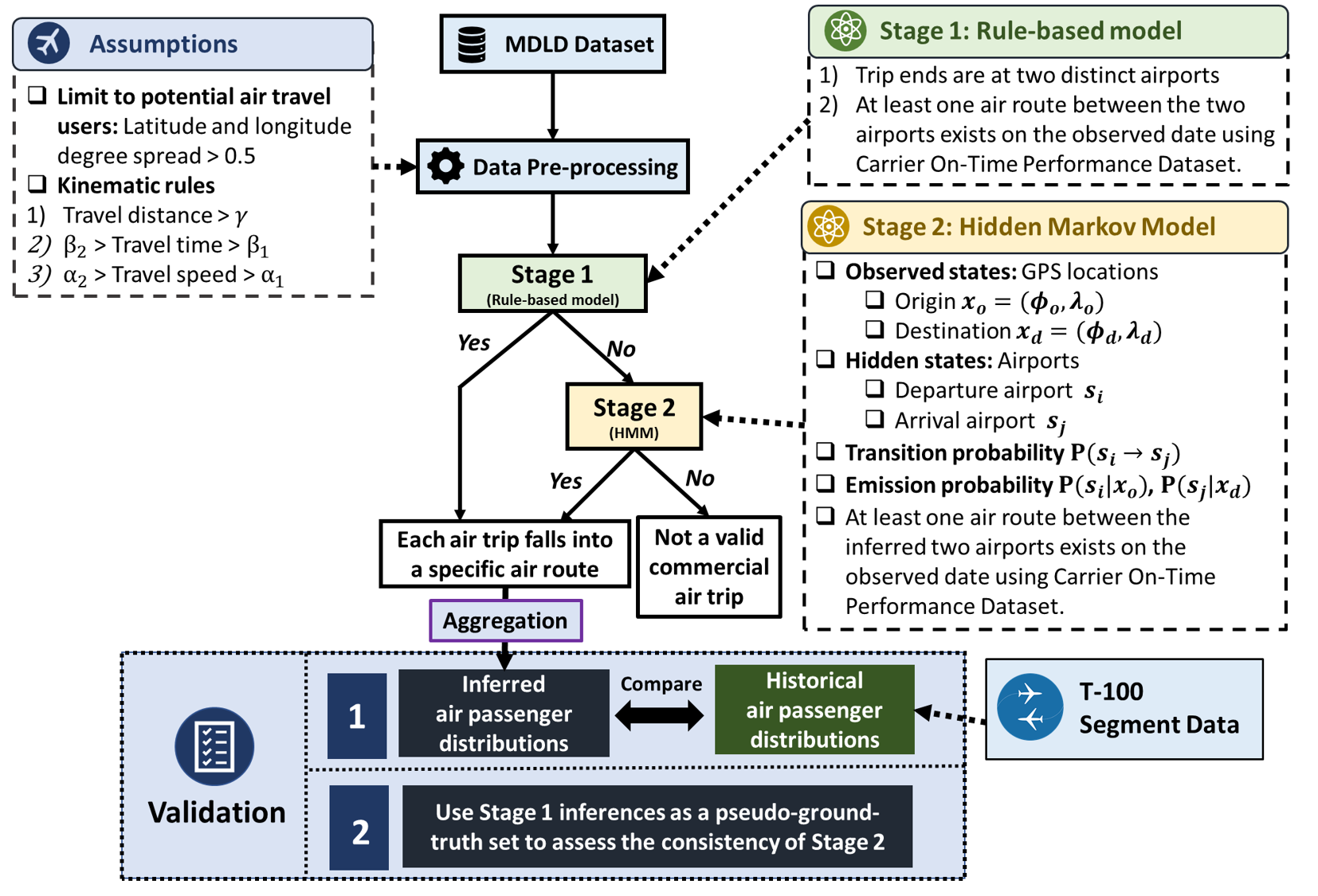}
 \caption{Overview of the proposed two-stage hybrid inference framework}
 \label{fig:framework-overview}
\end{figure}

\section{Methodology}\label{sec:data}

This study proposes a two-stage framework for inferring air trips, as shown in Figure~\ref{fig:framework-overview}. First, raw LBS data are filtered using a set of rules to retain devices associated with potential air travel. Stage~1 focuses on Type~1 air travelers, for whom sightings are explicitly observed at two distinct airports; a rule-based model is used to identify these high-confidence air trips. Devices that do not satisfy the Stage~1 criteria are processed in Stage~2, where a probabilistic model infers their most likely air routes.

\subsection{Assumptions}\label{subsec:preprocess}
To reduce the computational burden of processing nationwide LBS data while retaining devices associated with potential long-distance air travel, the following assumptions and filtering rules are applied:

- The latitude and longitude span of the sightings exceeds $0.5^\circ$, which is approximately 35 miles. This screen serves as a computational prefilter rather than a sufficient condition for classifying an air trip.

- Travel time is between 30 minutes ($\beta_{1}$) and 18 hours ($\beta_{2}$). The 18-hour upper bound retains long domestic U.S. air trips while allowing additional time for airport access, waiting, and egress. The 30-minute lower bound follows the air-trip classification threshold adopted in the Next Generation NHTS Passenger Origin-Destination Data \cite{NextGenNHTSOD}.
 
- Travel distance is greater than 50 miles ($\gamma$). This value follows the air-trip classification threshold adopted in the Next Generation NHTS Passenger Origin-Destination Data. \cite{NextGenNHTSOD}. 

- Travel speed is between 100 mph ($\alpha_{1}$) and 800 mph ($\alpha_{2}$). The 100-mph lower bound follows the air-trip classification threshold adopted in the Next Generation NHTS Passenger Origin-Destination Data, while the 800-mph upper bound reflects the upper range of commercial aircraft flight speeds \cite{NextGenNHTSOD,FAA2007}.

\begin{table}[!ht]
 \centering
 \caption{List of Key Notations}
 \label{tab:notation}
 \setlength{\tabcolsep}{4pt}
 \begin{tabular}{p{0.24\textwidth}p{0.68\textwidth}}
 \hline
 \multicolumn{2}{l}{\textbf{Variables}} \\
 \hline
 $u$ & An anonymized mobile-device user. \\
 $e_u$ & Candidate air-trip segment for user $u$. \\
 $x_o$, $x_d$ & Observed pre-transition and post-transition sighting locations. \\
 $t_o$, $t_d$ & Timestamps associated with $x_o$ and $x_d$, respectively. \\
 $O$ & Observed spatial state, $O=(x_o,x_d)$. \\
 $\phi$, $\lambda$ & Latitude and longitude of a location sighting. \\
 $s_i$, $s_j$, $s_k$ & Candidate airports indexed by $i$, $j$, and $k$. \\
 $S_A$ & Set of U.S. commercial airports. \\
 $S_o(u)$, $S_d(u)$ & Candidate origin- and destination-airport sets for device $u$. \\
 $\mathcal{R}(u)$ & Candidate directed route space, $S_o(u)\times S_d(u)$. \\
$r$, $r^*$ & Candidate air route and maximum-posterior inferred air route. \\
$d$ & Calendar date of the candidate air-trip segment. \\
$N$ & Number of top-ranked routes considered in the top-$N$ route coverage evaluation. \\
$f_{ij,d}$ & Number of scheduled direct services from airport $s_i$ to airport $s_j$ on date $d$. \\
 $F^{\mathrm{out}}_{i,d}$, $F^{\mathrm{in}}_{j,d}$ & Marginal outbound and inbound service availability for candidate airports. \\

 $\Delta T$ & Observed elapsed time between endpoint sightings, $t_d-t_o$. \\

 \hline
 \multicolumn{2}{l}{\textbf{Parameters}} \\
 \hline
 $\alpha_1$, $\alpha_2$ & Lower and upper thresholds for travel speed filtering. \\
 $\beta_1$, $\beta_2$ & Lower and upper thresholds for travel duration filtering. \\
 $\gamma$ & Minimum travel distance threshold. \\
 $\xi$ & Maximum search radius for generating airport candidates. \\
 $\alpha$ & Flight-supply weight in the airport utility function. \\
 $\beta$ & Distance-friction weight in the airport utility function. \\
 $\epsilon$ & Smoothing constant. \\
 $\sigma$ & Temporal tolerance in the Gaussian compatibility score. \\
$V_{\mathrm{c}}$ & Cruise speed of commercial passenger aircraft. \\
 $T_{\mathrm{b}}$ & Time buffer for access, waiting, and egress at an airport. \\
 \hline
 \end{tabular}
\end{table}

\subsection{Data Processing and Analytics}\label{subsec:data-sources}
This study uses a national-level mobile device location dataset for January 2022, with observations collected from multiple data vendors. The dataset contains spatial and temporal information for individual devices, including a randomly hashed device ID, latitude and longitude coordinates, timestamps, and provider-reported place names. Table~\ref{tab:sample-mdld} presents a sample of the LBS data. To address privacy concerns, noise has been applied to all entries in Table~\ref{tab:sample-mdld}; the sample is provided only to illustrate the type of information contained in LBS data.

Two Bureau of Transportation Statistics (BTS) aviation datasets are also used. The T-100 Domestic Segment data provide monthly segment-level traffic and capacity records for nonstop domestic services, including origin and destination airports, departures performed, seats, passengers, and related capacity measures. In this study, the January 2022 T-100 data are used as an external route-volume benchmark and are not used to calibrate the inference model \cite{BTS2022T100DomesticSegment}. The Carrier On-Time Performance dataset provides flight-level operating records for reporting carriers, including carrier, flight date, origin and destination airports, scheduled and actual departure and arrival times, elapsed time, cancellation status, and delay information. This dataset is used to construct daily route availability for air-route inference and route feasibility checks. \cite{BTS2022OnTimePerformance}.

\begin{table}[!t]
 \centering
 \caption{Sample of Mobile Device Location Data}
 \label{tab:sample-mdld}
 \scriptsize
 \begin{tabular}{llllp{0.38\textwidth}}
 \hline
 \textbf{ID} & \textbf{Timestamp} & \textbf{Latitude} & \textbf{Longitude} & \textbf{Place name} \\
 \hline
 Sfacd-243rg & 2022-01-07 00:10:09 & 26.069201 & -80.14125 & convenient store \\
 Sfacd-243rg & 2022-01-07 00:45:10 & 26.069323 & -80.14297 & Fort Lauderdale-Hollywood International Airport \\
 Sfacd-243rg & 2022-01-07 00:49:35 & 26.069323 & -80.14297 & Fort Lauderdale-Hollywood International Airport \\
 Sfacd-243rg & 2022-01-07 07:31:07 & 29.986477 & -95.34981 & Houston - George Bush Intercontinental Airport (IAH) \\
 Sfacd-243rg & 2022-01-07 07:33:14 & 29.986478 & -95.34982 & Houston - George Bush Intercontinental Airport (IAH) \\
  Sfacd-243rg & 2022-01-07 11:21:20 & 29.986563 & -95.34812 & chain coffee shop \\
 \hline
 \end{tabular}
\end{table}

\subsection{Stage 1: Rule-based Inference}\label{subsec:stage1}
 
Stage 1 is designed to retain Type-1 air travelers, as shown in Figure~\ref{fig:observation-regimes}. A valid Type-1 air traveler is flagged when: (1) the endpoints correspond to two distinct airport regions; and (2) at least one direct air route between the two airports was available on that sighting date. The output is intentionally conservative as this stage is designed to retain reliable airport-endpoint cases. 

\subsection{Stage 2: Probabilistic inference}\label{subsec:stage2}
Stage 2 is designed to address the Type-2 and Type-3 air travelers in Figure~\ref{fig:observation-regimes}, for which one or both endpoints are not directly sighted at an airport. In this stage, a Hidden Markov Model (HMM) is developed to infer their air routes.
The list of key notations used in this stage are presented in Table~\ref{tab:notation}.

\subsubsection{Observed State, Hidden State and Candidate Route Space}

A candidate air-trip segment for device $u$ is mathematically represented as:
\begin{equation}
 e_u = \left(x_o,t_o,x_d,t_d, \Delta{T} \right)
 \label{eq:air-trip-segment}
\end{equation}
where $x_o$ and $x_d$ denote the sighted locations immediately before and after a candidate long-distance transition, and $t_o$ and $t_d$ denote their corresponding timestamps. $\Delta{T}$ represents the elapsed time between these two sightings. 
The target output is a directed airport pair $(i,j)$, where $i$ is the inferred departure airport and $j$ is the inferred arrival airport. Throughout this paper, an \emph{air route} refers to a direct domestic segment between two airports; it does not represent a complete passenger ticket itinerary with one or more connections.

\noindent\textbf{Observed State:} The observed state $O$ refers to the spatial component of the candidate air-trip segment and is expressed as follows:
\begin{equation}
 O = (x_{o}, x_{d}),
 x_{o} = (\phi_o, \lambda_o),
 x_{d} = (\phi_d, \lambda_d),
\end{equation}
where $\phi$ and $\lambda$ represent latitude and longitude, respectively. This formulation treats the two endpoint sightings as the observable evidence used to infer the latent airport pair. The timestamps are not included in $O$ directly; instead, they are used later to evaluate temporal compatibility between the observed time gap and the expected travel time for each candidate air route.

\noindent\textbf{Hidden State:} The hidden states represent potential departure and arrival airports that are not directly observed in LBS data. Let $S_A$ denote the set of U.S. commercial airports. Given a maximum search radius $\xi$, the candidate departure-airport set $S_o$ is defined as follows:
\begin{equation}
 S_o(u) = \left\{s_i \in S_A \mid \operatorname{H}(x_o,s_i) \leq \xi\right\},
 \label{eq:candidate-origin-airports}
\end{equation}
Equation~\ref{eq:candidate-origin-airports} retains airports located within a distance $\xi$ of the pre-transition sighting $x_o$. These airports are treated as feasible origins because the traveler may have been observed before reaching the departure airport or shortly after leaving an airport.

Similarly, the candidate arrival-airport set $S_d$ is defined as follows:
\begin{equation}
 S_d(u) = \left\{s_j \in S_A \mid \operatorname{H}(x_d,s_j) \leq \xi\right\}.
 \label{eq:candidate-destination-airports}
\end{equation}
Equation~\ref{eq:candidate-destination-airports} retains airports located within a distance $\xi$ of the post-transition sighting $x_d$. These airports are treated as feasible destinations because the traveler may have been observed after leaving the arrival airport or near the airport after landing. The distance function $\operatorname{H}(\cdot, \cdot)$ is defined as the great-circle distance between two geographical coordinate points using the Haversine formula \cite{Sin84}. The \emph{candidate route space} for device $u$ is the Cartesian product of the endpoint candidate sets:
\begin{equation}
 \mathcal{R}(u)=S_o(u)\times S_d(u)
 =\left\{(s_i,s_j)\mid s_i\in S_o(u),\ s_j\in S_d(u)\right\}.
 \label{eq:candidate-route-space}
\end{equation}
Equation~\ref{eq:candidate-route-space} enumerates all directed airport pairs that can plausibly explain the observed long-distance transition. Each candidate route in $\mathcal{R}(u)$ is subsequently scored using emission probability, transition probability, and temporal compatibility. The goal of the proposed HMM is to identify the most likely air route within this candidate route space.

\subsubsection{Emission Probability}
Let $f_{ij,d}$ be the number of scheduled direct services from airport $s_i$ to airport $s_j$ on date $d$. The marginal service availability for a candidate origin and destination is defined as
\begin{align}
 F^{\mathrm{out}}_{i,d} &= \sum_{s_j\in S_d(u)} f_{ij,d}, \\
 F^{\mathrm{in}}_{j,d} &= \sum_{s_i\in S_o(u)} f_{ij,d}.
 \label{eq:marginal-service-frequency}
\end{align}

The utility score $U(s_i)$ for an air passenger choosing a specific airport $i$ is determined by both distance friction and total flight supply. Inspired by the gravity model concept, it is defined as follows:

\begin{equation}
U(s_i) = \frac{(F^{\mathrm{out}}_{i,d}+ \epsilon)^\alpha}{\mathrm{H}(x_{o}, s_{i})^\beta}
\label{eq:utility-score-origin}
\end{equation}

\begin{equation}
U(s_j) = \frac{(F^{\mathrm{in}}_{j,d}+ \epsilon)^\alpha}{\mathrm{H}(x_{d}, s_{j})^\beta}
\label{eq:utility-score-dest}
\end{equation}

\noindent where $\epsilon$ is a smoothing constant; $\alpha$ and $\beta$ are weight parameters.

Equations~\ref{eq:utility-score-origin} and~\ref{eq:utility-score-dest} encode the physical intuition that an airport is more likely to be the true endpoint when it is both close to the observed LBS data sighting and well connected by scheduled air service. In the numerator, $F^{\mathrm{out}}_{i,d}$ and $F^{\mathrm{in}}_{j,d}$ represent the service supply available from a candidate origin or to a candidate destination on date $d$; airports with more scheduled flights are assigned higher utility because they provide more feasible opportunities for the observed trip. The smoothing constant $\epsilon$ prevents airports with very low or zero marginal service counts from being eliminated too aggressively during scoring. In the denominator, the distance terms $\mathrm{H}(x_o,s_i)$ and $\mathrm{H}(x_d,s_j)$ represent spatial impedance: as the observed location becomes farther from a candidate airport, the utility decreases because the user is less likely to travel to the airport. The parameters $\alpha$ and $\beta$ control the relative influence of flight supply and distance friction, respectively.

Normalizing the utility scores within the candidate set yields the emission probability $P_{o}$ and $P_{d}$ for the origin and the destination:

\begin{equation}
P_{o}(S_{o} = s_i \mid x_{o}) = \frac{U(s_i)}{\sum_{k \in S_{o}} U(s_k)}
\label{eq:emission-probabilit-org}
\end{equation}

\begin{equation}
P_{d}(S_{d} = s_j \mid x_{d}) = \frac{U(s_j)}{\sum_{k \in S_{d}} U(s_k)}
\label{eq:emission-probabilit-dest}
\end{equation}

Equations~\ref{eq:emission-probabilit-org} and~\ref{eq:emission-probabilit-dest} convert the raw airport utility scores into probabilities over the candidate origin and destination sets. Physically, the emission probability measures how well a candidate airport can explain the observed endpoint sighting. For the origin side, $P_o(S_o=s_i\mid x_o)$ is large when airport $s_i$ has a high utility relative to all other candidate origins near $x_o$. By the same token, $P_d(S_d=s_j\mid x_d)$ is large when airport $s_j$ has a high utility relative to all other candidate destinations near $x_d$. The denominators normalize the utilities so that the probabilities across each candidate set sum to one, meaning that each airport is evaluated by its relative attractiveness rather than by its absolute utility alone. Therefore, a nearby airport with strong flight supply receives a higher emission probability, while a distant airport or one with limited service receives a lower probability.

\subsubsection{Transition Probability}

The transition probability aims to quantify the likelihood that an air passenger travels from airport $s_{i}$ to $s_{j}$. This is computed with the number of scheduled direct services from airport $s_i$ to airport $s_j$ on date $d$. Note that the denominator includes the air route frequency of all the airports in the nationwide airport set $S_{A}$.

\begin{equation}
P_{\mathrm{trans}}(s_j \mid s_i) = \frac{f_{ij,d}}{\sum_{s_k \in S_A,\; k \neq i} f_{ik,d}}
\end{equation}

\subsubsection{Gaussian Temporal Compatibility}
To reduce false positives caused by anomalous GPS sightings or long-distance ground movements, the model penalizes inconsistencies between the LBS data sighting time gap and the expected elapsed time for a candidate air route. Let the sighting time gap be defined as:
\begin{equation}
 \Delta T=t_d-t_o
 \label{eq:observed-time-gap}
\end{equation}
Equation~\ref{eq:expected-time} defines the expected elapsed time:
\begin{equation}
 T_{\mathrm{e}}(s_i,s_j)=
 \frac{\operatorname{H}(s_i,s_j)}{V_{\mathrm{c}}}
 +T_{\mathrm{b}},
 \label{eq:expected-time}
\end{equation}
where $V_{\mathrm{c}}$ is the cruise speed of commercial passenger aircraft and $T_{\mathrm{b}}$ represents the aggregate time not captured by air travel alone, including airport access, waiting, and egress. The temporal compatibility score is defined as:
\begin{equation}
 P_{\mathrm{temp}}\left(\Delta T\mid s_i,s_j\right)=
 \exp\!\left[-\frac{\left\{\Delta T-T_{\mathrm{e}}(s_i,s_j)\right\}^2}{2\sigma^2}\right],
 \label{eq:temporal-compatibility}
\end{equation}
where $\sigma$ controls the tolerance for deviations from the expected elapsed time. 

Equation~\ref{eq:temporal-compatibility} measures whether the observed time gap between two LBS data sightings is physically consistent with the travel time implied by a candidate airport pair. When $\Delta T$ is close to the expected elapsed time $T_{\mathrm{e}}(s_i,s_j)$, the exponent is near zero and $P_{\mathrm{temp}}$ approaches one, indicating that the route is temporally plausible. As the observed time gap becomes much shorter or much longer than the expected elapsed time, the squared difference increases, and the compatibility score decreases exponentially. Thus, $\sigma$ determines how strictly the model applies this temporal filter: a smaller value imposes stronger penalties for timing mismatches, whereas a larger value allows greater uncertainty in non-flight time components, such as airport access, waiting, flight delays, and egress.

\subsubsection{Joint Posterior and Route Assignment}
For each candidate route $r=(s_i,s_j)\in\mathcal{R}(u)$, by the Chain rule of probability, it can be mathematically expressed as: 

\begin{equation}
P(s_i, s_j, x_o, x_d, \Delta{T}) = P(s_i) \cdot P(x_o|s_i) \cdot P(s_j|s_i, x_o) \cdot P(x_d|s_i, x_o, s_j) \cdot P(\Delta{T}|s_i, x_o, s_j, x_d)
\label{eq:HMM_joint_prob}
\end{equation}

Based on the assumptions of Markov property and output independence, Equation ~\ref{eq:HMM_joint_prob} can be further expressed as:

\begin{equation}
P(s_i) \cdot P(x_o|s_i) \cdot P(s_j|s_i) \cdot P(x_d|s_j) \cdot P(\Delta{T}|s_i, x_o, s_j, x_d)
\end{equation}

Assuming $P_{temp}(\Delta{T}|s_i, x_o, s_j, x_d)$ only depends on $(s_i, s_j)$, it is equivalent to:

\begin{equation}
 \widetilde{P}(r\mid O)\propto
 P_o(s_i\mid x_o)\,
 P_d(s_j\mid x_d)\,
 P_{\mathrm{trans}}(s_j\mid s_i)\,
 P_{\mathrm{temp}}\left(\Delta T\mid s_i,s_j\right).
 \label{eq:unnormalized-posterior}
\end{equation}

Equation~\ref{eq:unnormalized-posterior} represents the combined evidence that a candidate air route explains the observed LBS data segment $O$. The origin and destination emission terms measure whether the two endpoint sightings are spatially consistent with airports $s_i$ and $s_j$; the transition term reflects the likelihood of travel from $s_i$ to $s_j$ based on scheduled flight frequency; and the temporal compatibility term evaluates whether the observed sighting time gap is consistent with the expected elapsed time for that route. A candidate route receives a high posterior score only when all components provide strong support. 

Finally, the inferred air route is the maximum posterior candidate,
\begin{equation}
 r^*=\arg\max_{r\in\mathcal{R}(u)}\widetilde{P}(r\mid O).
 \label{eq:maximum-posterior-route}
\end{equation}

\begin{table}[!htbp]
 \centering
 \caption{Parameter settings used in the probabilistic inference model}
 \label{tab:model-parameter-settings}
 \small
 \begin{tabular}{p{0.40\textwidth}p{0.25\textwidth}p{0.20\textwidth}}
 \hline
 \textbf{Parameter} & \textbf{Symbol} & \textbf{Value} \\
 \hline
 Maximum search radius & $\xi$ & 50 miles \\
 Origin/destination supply weight & $\alpha$ & 1.0 \\
 Distance friction weight & $\beta$ & 1.0 \\
 Temporal tolerance & $\sigma$ & 120 min \\
 Cruise speed & $V_{\mathrm{c}}$ & 500 mph \\
 Time buffer & $T_{\mathrm{b}}$ & 120 min \\
 Smoothing constant & $\epsilon$ & 1.0 \\
 \hline
 \end{tabular}
\end{table}

Table~\ref{tab:model-parameter-settings} summarizes the preset parameters used to implement the Stage~2 model. These values are held fixed across all experiments to provide a consistent baseline for evaluating the two-stage framework.

\section{Experimental Design}\label{sec:evaluation}

Two experiments are designed to validate and evaluate the proposed framework.

\subsection{Aggregate Route-Distribution Validation}\label{subsec:external-validation}
The aggregate external validation compares the inferred route distribution with T-100 Domestic Segment passenger volumes for the same month. For each airport-pair route $(i,j)$, the inferred share and the observed T-100 route share are computed over the same geographic and temporal scope. The analysis reports three main metrics:  Spearman correlation, mean absolute error, and Jensen--Shannon divergence. Note that the T-100 data are not used to calibrate model parameters and are reserved only for external aggregate validation.

Let $\widehat{p}_r$ denote the inferred share for route $r\in\mathcal{R}$, $p_r$ denote the corresponding T-100 route share, and $\widehat{R}_r$ and $R_r$ denote their ranks. Spearman correlation ($\rho_s$) measures rank-order similarity between the inferred and observed route shares; values closer to one indicate that the model better preserves the relative ordering of high- and low-volume air routes. Mean absolute error (MAE) measures the average absolute difference between inferred and observed route shares. Jensen--Shannon divergence (JSD) compares the full distributions of inferred and observed route shares, with smaller values indicating greater similarity between the two distributions.
These metrics are computed as:
\begin{align}
 \rho_s &=
 \frac{\sum_{r\in\mathcal{R}}(\widehat{R}_r-\overline{\widehat{R}})(R_r-\overline{R})}
 {\sqrt{\sum_{r\in\mathcal{R}}(\widehat{R}_r-\overline{\widehat{R}})^2}
  \sqrt{\sum_{r\in\mathcal{R}}(R_r-\overline{R})^2}},
 \label{eq:spearman-correlation}\\
 \mathrm{MAE} &=
 \frac{1}{|\mathcal{R}|}\sum_{r\in\mathcal{R}}\left|\widehat{p}_r-p_r\right|,
 \label{eq:share-mae}\\
 \mathrm{JSD}(\widehat{\mathbf{p}},\mathbf{p}) &=
 \frac{1}{2}D_{\mathrm{KL}}(\widehat{\mathbf{p}}\parallel\mathbf{m})+
 \frac{1}{2}D_{\mathrm{KL}}(\mathbf{p}\parallel\mathbf{m}),
 \quad
 \mathbf{m}=\frac{1}{2}(\widehat{\mathbf{p}}+\mathbf{p}),
 \label{eq:jensen-shannon-divergence}
\end{align}
where $D_{\mathrm{KL}}(\mathbf{a}\parallel\mathbf{b})=\sum_{r\in\mathcal{R}}a_r\log(a_r/b_r)$ is the Kullback--Leibler divergence.


\subsection{Internal Consistency Against the Strict Reference Set}\label{subsec:consistency}
This experiment evaluates the consistency of Stage~2 by testing whether it can reproduce the air-route inferences identified by Stage~1. A unique combination of device identifier, flight date, origin airport, and destination airport from the Stage~1 results is used as the unit of analysis and serves as a pseudo-ground-truth reference. The same preprocessed LBS data are then input directly into Stage~2, without first passing through Stage~1, to generate inferred air-route results. The analysis reports the coverage of Stage~1 reference trips by Stage~2.

\section{Results}\label{sec:results}

The proposed two-stage framework was implemented and evaluated using the nationwide mobile device location dataset described in the methodology. For January 2022, the raw dataset contained nearly 10 million anonymized mobile-device users and approximately 131 million location sightings, providing extensive spatial and temporal coverage of travel activity across the United States. The large scale of the dataset allowed the framework to be tested under realistic conditions involving sparse, irregular, and incomplete airport observations. Data preprocessing included filtering users based on geographic movement, travel distance, elapsed time, and implied speed, after which candidate trips were processed using the rule-based and probabilistic inference stages. All preprocessing, route inference, aggregation, and validation experiments were conducted on a high-performance workstation equipped with 72 virtual central processing units and 188 GiB of memory. These computational resources were necessary to efficiently process the nationwide dataset, generate candidate airport pairs, evaluate route probabilities, and compare the resulting route distributions with external and internal validation benchmarks.

\subsection{Aggregate Route-Distribution Validation}\label{subsec:results-aggregate}
Tables~\ref{tab:route-level-detection-performance}--\ref{tab:route-share-distribution-metrics} summarize the aggregate comparison between the inferred air routes and the T-100 Domestic Segment route set. The T-100 data reports 9,339 directed domestic air routes. Stage~1 identifies 995 unique routes, corresponding to 10.65\% route coverage. By contrast, the proposed framework identifies 4,726 unique routes, increasing route coverage to 50.60\%. This improvement indicates that many operationally active routes are missing from airport-sighted cases alone and can be supplemented by the inference from semi-sighted or latent air trips.

\newpage
\begin{table}[htbp]
 \centering
 \caption{Route-level detection performance against T-100 Domestic Segment data}
 \label{tab:route-level-detection-performance}
 \small
 \begin{tabular}{p{0.24\textwidth}p{0.22\textwidth}p{0.22\textwidth}p{0.18\textwidth}}
 \hline
 \textbf{Model} & \textbf{Identified unique air routes} & \textbf{Actual air routes from T-100 data} & \textbf{Coverage} \\
 \hline
 Stage 1 only & 995 & 9,339 & 10.65\% \\
 Proposed framework & 4,726 & 9,339 & 50.60\% \\
 \hline
 \end{tabular}
\end{table}
\vspace{-0.75em}

\begin{table}[htbp]
 \centering
 \caption{Top-$N$ route coverage against T-100 Domestic Segment data}
 \label{tab:top-n-route-coverage}
 \small
 \begin{tabular}{p{0.30\textwidth}p{0.27\textwidth}p{0.27\textwidth}}
 \hline
 \textbf{Air route group} & \textbf{Stage 1} & \textbf{Proposed framework} \\
 \hline
 Top 20 routes & 19 / 20 (95.00\%) & 20 / 20 (100.00\%) \\
 Top 50 routes & 46 / 50 (92.00\%) & 48 / 50 (96.00\%) \\
 Top 100 routes & 88 / 100 (88.00\%) & 98 / 100 (98.00\%) \\
 Top 500 routes & 357 / 500 (71.40\%) & 486 / 500 (97.20\%) \\
 Top 1,000 routes & 556 / 1,000 (55.60\%) & 973 / 1,000 (97.40\%) \\
 All recovered air routes & 995 / 9,339 (10.65\%) & 4,726 / 9,339 (50.60\%) \\
 \hline
 \end{tabular}
\end{table}
\vspace{-0.75em}

\begin{table}[htbp]
 \centering
 \caption{Route-share distribution metrics against T-100 Domestic Segment data}
 \label{tab:route-share-distribution-metrics}
 \small
 \begin{tabular}{p{0.36\textwidth}p{0.20\textwidth}p{0.24\textwidth}}
 \hline
 \textbf{Metric} & \textbf{Stage 1} & \textbf{Proposed framework} \\
 \hline
 Spearman correlation & 0.4490 & 0.7630 \\
 Share MAE & 0.0123\% & 0.0117\% \\
 JS divergence & 0.2808 & 0.2263 \\
 \hline
 \end{tabular}
\end{table}

The top-route comparison further shows that the proposed framework substantially improves coverage of high-volume routes. For the top-20 air routes, Stage-1 recovers 19 routes, while the proposed framework recovers all 20. The difference becomes larger as the route set expands: coverage increases from 88\% to 98\% for the top 100 routes, from 71.40\% to 97.20\% for the top 500 routes, and from 55.60\% to 97.40\% for the top 1,000 routes. These results suggest that the proposed framework preserves the major structure of the domestic air-route network more completely than the rule-based method (i.e., Stage 1).

\begin{figure}[!t]
 \centering
 \includegraphics[width=\textwidth]{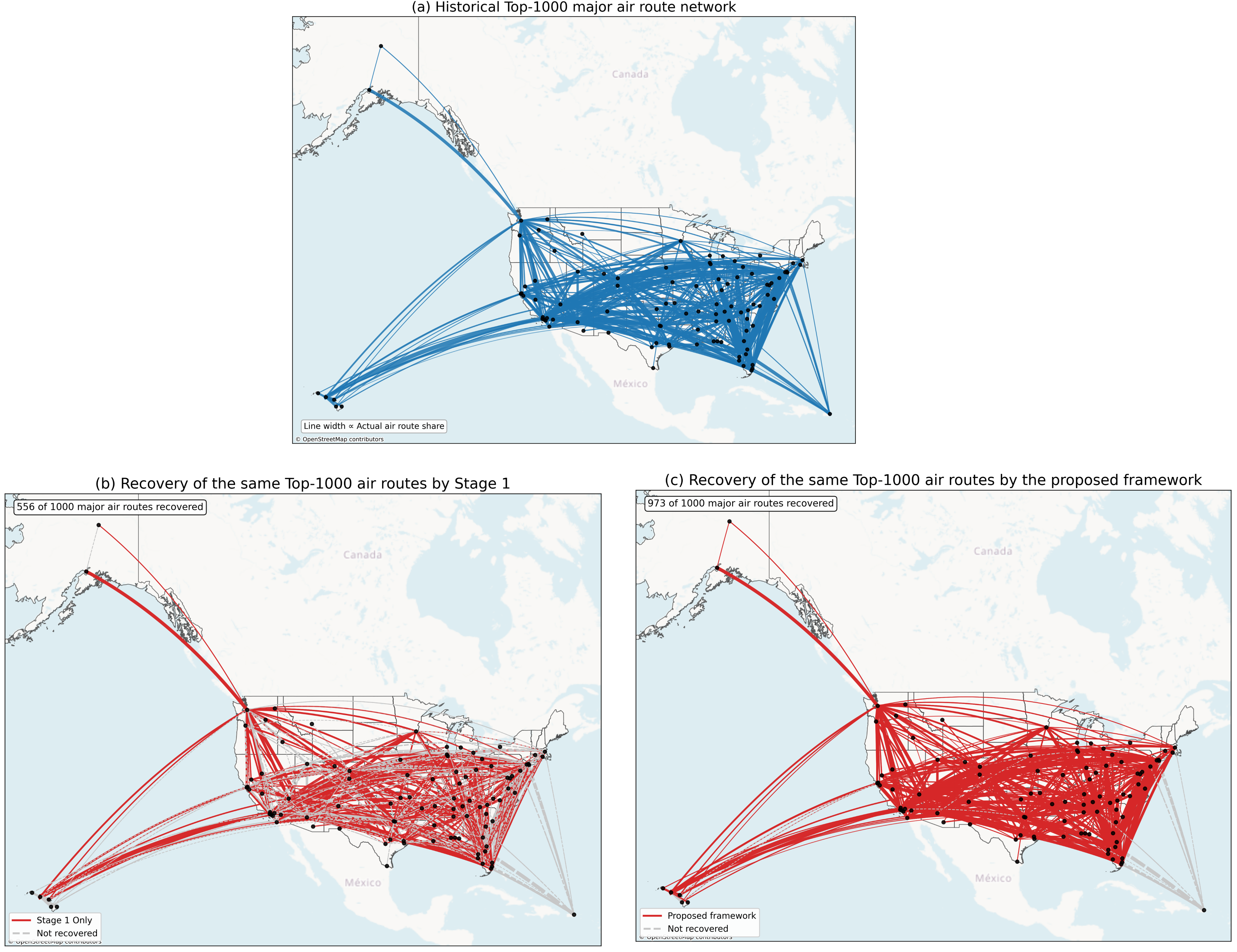}
 \caption{Recovery of the historical top-1,000 domestic air-route network}
 \label{fig:top1000-network-recovery}
\end{figure}

Figure~\ref{fig:top1000-network-recovery} visualizes this pattern for the top 1,000 air routes. The historical network is dense across the continental U.S., with additional long-haul routes linking Alaska, Hawaii, and major mainland hubs. Stage~1 recovers 556 of these 1,000 routes and leaves visible gaps in the recovered network. The proposed framework recovers 973 of the top 1,000 routes, producing a network that more closely resembles the historical air route structure.

Distribution-level metrics provide a more nuanced interpretation beyond route-count recovery. The proposed framework improves rank-order agreement with T-100 route shares, increasing Spearman correlation from 0.4490 to 0.7630. This result indicates that the inferred routes are not only more numerous but also more consistently ordered with respect to their relative importance in the T-100 data. The framework also slightly reduces the share mean absolute error from 0.0123\% to 0.0117\%, suggesting an improvement in the average route-share discrepancy across the route set. In addition, Jensen--Shannon divergence (JS divergence) decreases from 0.2808 to 0.2263, indicating that the overall inferred route-share distribution is closer to the T-100 distribution. In summary, these metrics suggest that the proposed framework improves both network coverage and distributional similarity relative to Stage~1.

\subsection{Internal Consistency Against the Strict Reference Set}\label{subsec:results-feasibility}
Model consistency is evaluated by comparing the proposed probabilistic inference model with the high-confidence reference trips identified by Stage~1. In this validation, a unique air trip is defined by the combination of user identifier, flight date, origin airport, and destination airport. Stage~1 identifies 2,073 unique airport-sighted air trips, which serve as the pseudo-ground truth reference set. Stage~2 identifies 305,563 unique air trips in total, reflecting its broader inference scope across airport-sighted, semi-sighted, and latent air trips.

\begin{table}[!t]
 \centering
 \caption{Internal consistency of Stage~2 against Stage~1 pseudo-ground-truth reference set}
 \label{tab:model-consistency}
 \small
 \begin{tabular}{p{0.56\textwidth}p{0.30\textwidth}}
 \hline
 \textbf{Metric} & \textbf{Value} \\
 \hline
 Stage~1 unique air-trip count & 2,073 \\
 Stage~2 unique air-trip count & 305,563 \\
 Overlapped unique air-trip count & 1,911 \\
 Stage~1 trips missed by Stage~2 & 162 \\
 Stage~2-only trips & 303,490 \\
 Coverage of Stage~1 trips by Stage~2 & 92.2\% \\
 \hline
 \end{tabular}
\end{table}

As shown in Table~\ref{tab:model-consistency}, Stage~2 recovers 1,911 of the 2,073 Stage~1 reference trips, corresponding to 92.2\% coverage of the air routes inferred from Stage~1. The Stage~1 air-route inferences were derived from airport place names, whereas Stage~2 did not use place names and instead relied on endpoint coordinates, timestamps, and probabilistic route scores. The 92.2\% overlap therefore indicates substantial agreement between two distinct inference mechanisms for airport-observed trips. This result shows that the Stage~2 model can reproduce air route inferences when airport endpoints are directly observed.

At the same time, Stage~2 produces 305,563 additional unique trips that are not captured by Stage~1. These Stage~2-only trips represent the added coverage obtained by extending inference beyond airport-sighted cases to trips with one missing airport endpoint or no explicit airport endpoint sightings. The 162 Stage~1 trips that are not recovered by Stage~2 suggest that some airport-sighted cases are excluded by the constraints in the second stage; these cases provide useful targets for future threshold calibration and error analysis.

\section{Conclusion}\label{sec:conclusion}
This study proposes a two-stage framework for inferring U.S. domestic air trips between airport pairs from massive LBS data. By distinguishing strict airport-sighted cases from partially sighted and latent air trips, the framework aligns the inference approach with the available endpoint sighting evidence. Stage~1 identifies airport-sighted trips using heuristic kinematic rules, whereas Stage~2 treats the airport pair as a latent variable and assigns probabilities to feasible air routes using spatial, temporal, and air service constraints.

The experimental results show that the proposed framework substantially improves domestic air-route network recovery relative to the strict rule-based stage alone. The framework also improves rank-order agreement with observed air-route shares. The internal consistency evaluation further shows that Stage~2 recovers 92.2\% of the Stage~1 pseudo-ground-truth trips, suggesting that the probabilistic inference stage remains broadly consistent with high-confidence airport-sighted detections while extending coverage to less directly observed trips. 

Future extensions of this research could support broader analyses of aviation demand, including the characterization of airport access and egress patterns, passenger arrival times at airports, and travel behaviors after arrival. This study also highlights privacy concerns associated with LBS data: if inadequately protected, detailed location traces could be used to infer sensitive individual mobility patterns through frameworks similar to the one proposed here.

Several limitations remain. First, the probabilistic model parameters are preset rather than fully calibrated, which may affect route-assignment accuracy and the estimated distribution of inferred airport-pair trips. Sensitivity analyses of the model parameters should therefore be conducted in future work. Second, the empirical analysis is limited to a single study month; applying the framework to additional months would help evaluate the robustness of model performance across temporal and seasonal conditions. Third, the internal consistency assessment evaluates airport-observed reference trips and does not directly quantify route reconstruction accuracy when one or both airport endpoint sightings are unavailable. Future work can use controlled masking experiments to evaluate inference performance under these more challenging conditions.

\section{Authors' Note}
The authors acknowledge that the large language model ChatGPT 5.5 was used to support grammar editing and code troubleshooting during manuscript preparation. The authors recognize the limitations of language models and have rigorously verified the accuracy and the validity of the resulting code, text, and analyses.

\section*{AUTHOR CONTRIBUTIONS}
The authors confirm contribution to the paper as follows: 
study conception and design: Hsuan-Chih Wang and Xianfeng Terry Yang, data analysis and interpretation of results: Hsuan-Chih Wang
draft manuscript preparation: Hsuan-Chih Wang and Xianfeng Terry Yang. All authors reviewed the results and approved the final version of the manuscript.

\section*{DECLARATION OF CONFLICTING INTERESTS}
The authors declare no potential conflicts of interest with respect to the research, authorship, and publication of this article. 


\bibliographystyle{trb}
\bibliography{surveyed_literature}

\end{document}